\documentclass[10pt,twocolumn,letterpaper]{article}

\usepackage[pagenumbers]{cvpr} 
\usepackage[utf8]{inputenc}
\usepackage[T1]{fontenc}
\usepackage{times}
\usepackage{epsfig}
\usepackage{graphicx}
\usepackage{amsmath}
\usepackage{amssymb}

\usepackage{graphicx}
\usepackage{amsmath}
\usepackage{amssymb}
\usepackage{booktabs}
\usepackage{multirow}
\usepackage{pifont}
\usepackage{color, colortbl}
\usepackage{xcolor}
\usepackage{tikz}

\newcommand{\cmark}{\ding{51}}
\newcommand{\xmark}{\ding{55}}

\definecolor{gold}{HTML}{BD820B}
\definecolor{silver}{HTML}{909090}
\definecolor{bronze}{HTML}{9A5F26}

\newcommand*\circledd[1]{\tikz[baseline=(char.base)]{
            \node[shape=circle,draw,inner sep=0.15pt] (char) {#1};}}      

\newcommand{\first}[1]{%
    {#1\raisebox{0.8pt}{\footnotesize \color{gold} \circledd{1}}}%
}
\newcommand{\second}[1]{%
    {#1\raisebox{0.8pt}{\footnotesize \color{silver} \circledd{2}}}%
}
\newcommand{\third}[1]{%
    {#1\raisebox{0.8pt}{\footnotesize \color{bronze} \circledd{3}}}%
}

\usepackage[pagebackref,breaklinks,colorlinks]{hyperref}

\usepackage[capitalize]{cleveref}
\crefname{section}{Sec.}{Secs.}
\Crefname{section}{Section}{Sections}
\Crefname{table}{Table}{Tables}
\crefname{table}{Tab.}{Tabs.}

\def\iccvPaperID{12275} 

\begin{document}

\title{A Low-Shot Object Counting Network With Iterative Prototype Adaptation}

\author{Nikola Đukić, Alan Lukežič, Vitjan Zavrtanik, Matej Kristan \\
{\small Faculty of Computer and Information Science, University of Ljubljana, Slovenia} \\
{\tt\small nikola.m.djukic@gmail.com, \{alan.lukezic, vitjan.zavrtanik, matej.kristan\}@fri.uni-lj.si}
\vspace{-0.5cm}
}
\maketitle


\begin{abstract}
We consider low-shot counting of arbitrary semantic categories in the image using only few annotated exemplars (few-shot) or no exemplars (no-shot). The standard few-shot pipeline follows extraction of appearance queries from exemplars and matching them with image features to infer the object counts. 
Existing methods extract queries by feature pooling which neglects the shape information (e.g., size and aspect) and leads to a reduced object localization accuracy and count estimates.

We propose a Low-shot Object Counting network with iterative prototype Adaptation (LOCA).
Our main contribution is the new object prototype extraction module, which iteratively fuses the exemplar shape and appearance information with image features.
The module is easily adapted to zero-shot scenarios, enabling LOCA to cover the entire spectrum of low-shot counting problems.
LOCA outperforms all recent state-of-the-art methods on FSC147 benchmark by 20-30\% in RMSE on one-shot and few-shot and achieves state-of-the-art on zero-shot scenarios, while demonstrating better generalization capabilities.
The code and models are available here: \url{https://github.com/djukicn/loca}.
\end{abstract}

\section{Introduction}

Object counting considers estimation of the number of specific objects in the image. Solutions based on object detectors have been extensively explored for categories such as people~\cite{crowdcounting, crowdcounting2}, cars~\cite{cars, carpk} or animal species~\cite{animals, polyps}. However, 
these methods require huge annotated training datasets and are not applicable to counting new, previously unobserved, classes with potentially only few annotations. The latter problem is explored by low-shot counting, which encompasses few-shot and zero-shot counting. Few-shot counters count all present objects of some class with only few of them annotated by bounding boxes (exemplars), while zero-shot counters consider counting the most frequent class without annotations.

\begin{figure}[h!]
    \centering
    \includegraphics[width=\linewidth]{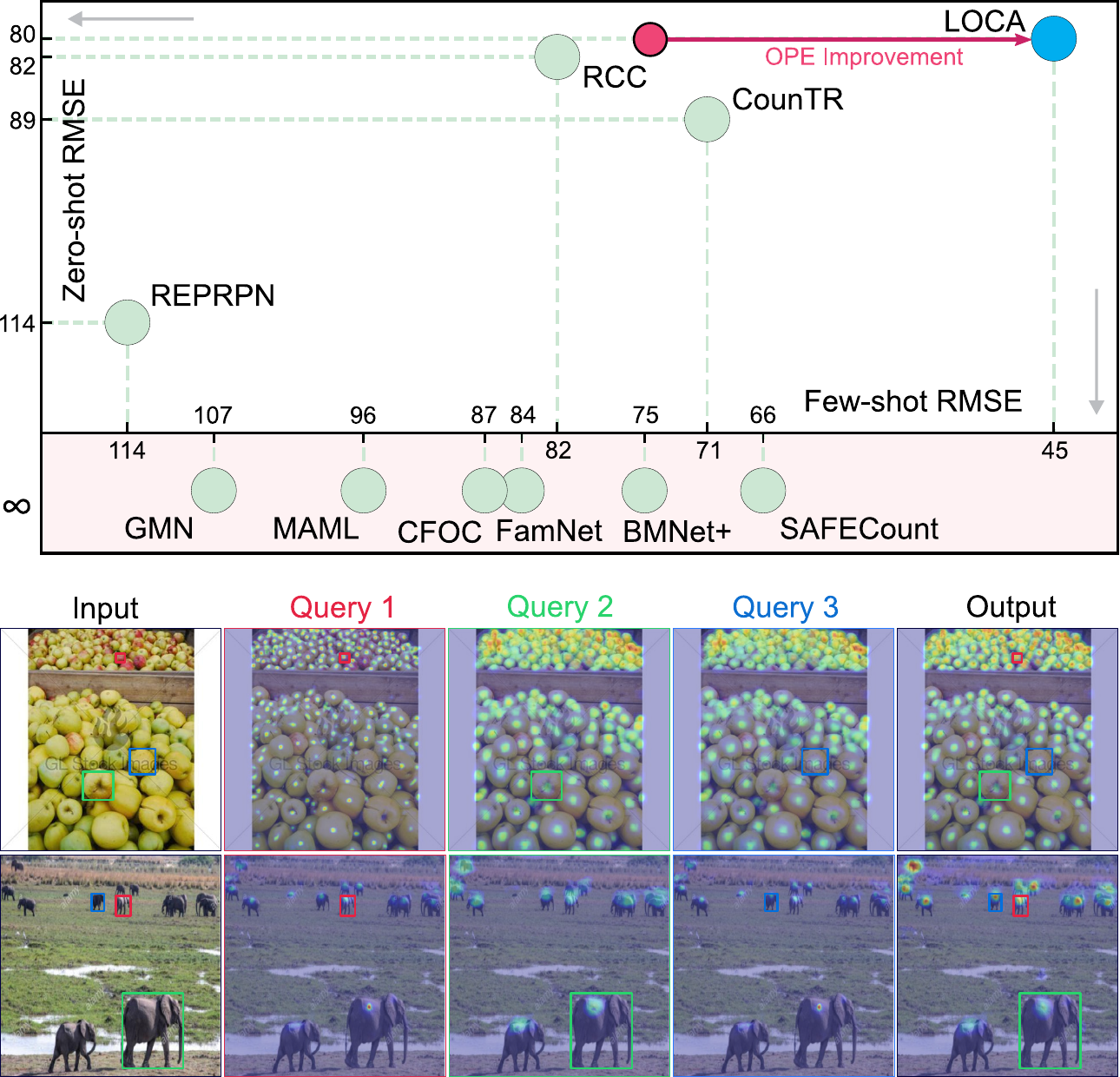}
    \caption{ 
    LOCA injects shape and appearance information into object queries to precisely count objects of various sizes in densely and sparsely populated scenarios. It also extends to a zero-shot scenario and achieves excellent localization and count errors across the entire low-shot spectrum.
    }
    \label{fig:figure1}
\end{figure}

Few-shot counters have recently gained momentum with the emergence of a challenging dataset~\cite{famnet} and follow a common pipeline~\cite{gmn,famnet,laonet,bmnet,safecount}. Image and exemplar features are extracted into object prototypes, which are matched to the image by correlation. Finally, the obtained intermediate image representation is regressed into a 2D object density map, whose values sum to the object count estimate.
The methods primarily differ in the intermediate image representation construction method, which is based either on 
Siamese similarity~\cite{gmn,famnet},
cross-attention~\cite{countr,laonet} 
or feature and similarity fusion~\cite{bmnet,safecount}.
While receiving much less attention, zero-shot counters follow a similar principle, but either identify possible exemplars by majority vote from region proposals~\cite{repprncount} or implicitly by attention modules~\cite{hobley}.

All few-shot counters 
construct object prototypes by pooling image features extracted from the exemplars into fixed-sized correlation filters.
The prototypes thus fail to encode the object shape information (i.e., width, height and aspect), resulting in a reduced accuracy of the density map. Recent works have shown that this information loss can be partially addressed by complex architectures for learning a nonlinear similarity function~\cite{bmnet}. Nevertheless, we argue that a much simpler counting architecture can be used instead, by explicitly addressing the exemplar shape and by applying an appropriate object prototype adaptation method. 

We propose a \underline{L}ow-shot \underline{O}bject \underline{C}ounting network with iterative prototype \underline{A}daptation (LOCA). 
Our main contribution is the new object prototype extraction module, which separately extracts the exemplar shape and appearance queries.
The shape queries are gradually adapted into object prototypes by considering the exemplar appearance as well as the appearance of non-annotated objects, obtaining excellent localization properties and leading to highly accurate counts (Figure~\ref{fig:figure1}).
To the best of our knowledge, LOCA is the first low-shot counting method that explicitly uses exemplars shape information for counting.
In contrast to most works~\cite{bmnet,famnet,cfocnet,safecount}, LOCA does not attempt to transfer exemplar appearance onto image features, but rather constructs strong prototypes that generalize across the image-level intra-class appearance. 

LOCA outperforms all state-of-the-art (in many cases more complicated methods) on the recent FSC147 benchmark~\cite{famnet}. 
On the standard few-shot setup it achieves $\sim$30\% relative performance gains, on one-shot setup even outperforms methods specifically designed for this setup, achieves state-of-the-art on zero-shot counting.
In addition, LOCA demonstrates excellent cross-dataset generalization on the car counting dataset CARPK~\cite{carpk}.

\section{Related work}

Historically, object counting has been addressed by class-specific detectors for people~\cite{crowdcounting, crowdcounting2}, cars~\cite{cars, carpk} and animals~\cite{animals}, but these methods do not cope well with extremely crowded scenes. In a jellyfish polyp counting scenario,~\cite{polyps} thus proposed to segment the image and interpret the segmentation as a collection of circular objects.
Alternatively,~\cite{crowdcounting, regression1} framed counting as a regression of object density map, whose summation predicts the number of objects. 
A major drawback of these methods is that they require large annotated training datasets for each object class, which is often an unrealistic requirement.

In response, class-agnostic counters have been explored, that specialize to the object category at test-time using only a few user-provided object exemplars.
An early representative~\cite{gmn} proposed a two-stream Generic Matching Network, that extracts the image and exemplar object features, concatenates them and regresses the representation into the final density map. 
CFOCNet~\cite{cfocnet} noted that a mere concatenation leads to unreliable localization and proposed a Siamese correlation network inspired by the tracking literature~\cite{siamfc} to improve the localization and counts.
Ranjan et al.~\cite{famnet} proposed a further improvement of correlation robustness by test-time Siamese backbone adaptation.
Shi et al.~\cite{bmnet} proposed an alternative approach for jointly learning the representation as well as a nonlinear similarity metric for improved localization and applied self-attention to reduce the within-class appearance variability in the test image.
You et al.~\cite{safecount} combined the similarity map with the image features before applying location regression to improve count accuracy and proposed a learnable similarity metric to guide the fusion of exemplar and image features.
Liu et al.~\cite{countr} adopted a vision transformer~\cite{vit} for image feature extraction and a convolutional encoder to extract the exemplars. Cross-attention is  used to fuse image and exemplar features and a convolutional decoder regresses the density map.
Recently, few-shot counting has been extended to few-shot detection~\cite{countingdetr} by adopting the transformer-based object detector~\cite{anchor_detr} to predict also the object bounding box in addition to location.

While most works addressed situations with several (typically three) exemplars available, only few recent works considered reducing this number.
Lin et al.~\cite{laonet} proposed a counting method that requires only a single exemplar. 
Their method is based on a transformer architecture and formulates correlation between image and exemplar features by several self- and cross-attention blocks. 
An extreme case of zero-shot counting~\cite{repprncount, hobley} has been explored as well.
Ranjan and Hoai~\cite{repprncount} proposed RepRPN-Counter, which combines a region proposal network~\cite{fasterrcnn} that also predicts a repetition score of each proposal. 
Proposals with the highest repetition scores are used as exemplars and sent through FamNet~\cite{famnet} to predict multiple density maps.
On the other hand, Hobley and Prisacariu~\cite{hobley} developed a weakly-supervised method that implicitly identifies object category most likely to be counted and predicts a density map for that category. Vision transformer with a unsupervised training stage~\cite{countr} has also shown success in zero-shot counting.

\section{A low-shot prototype adaptation counter}\label{sec:method}

\begin{figure*}[t]
    \centering
    \includegraphics[width=1.0\linewidth]{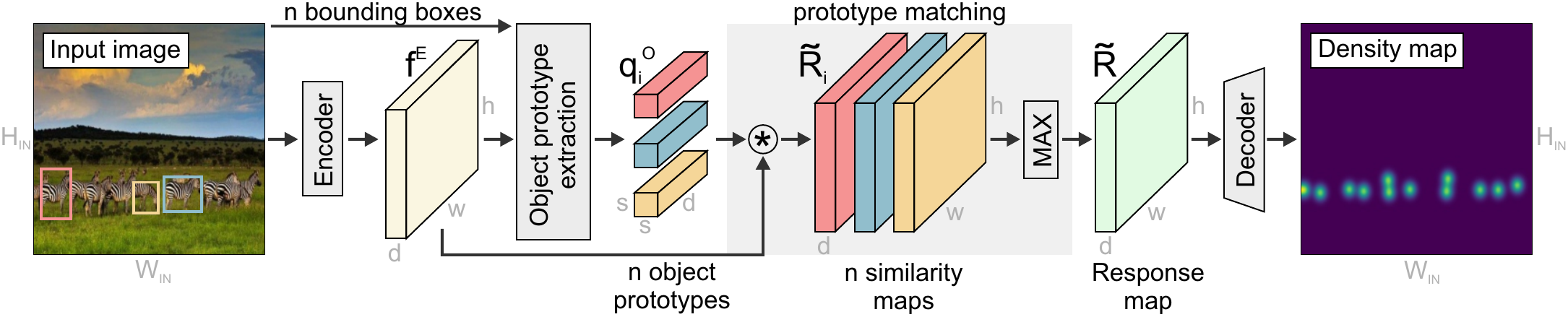}
    \caption{The LOCA architecture.
    Input image is encoded into features $\mathbf{f}^{E}$, which are depth-wise correlated (*) by $n$ object queries predicted by the object prototype extraction module. 
    The response map $\tilde{\mathbf{R}}$ is obtained by computing per-element maximum of $n$ similarity maps $\tilde{\mathbf{R}}_i$ and then upsampled by decoder to the final density map.
    }
    \label{fig:architecture}
\end{figure*}

Without loss of generality, we present our low-shot counting method LOCA in the context of few-shot counting.
Given an input image $\mathbf{I} \in \mathbb{R}^{H_0 \times W_0 \times 3}$ and a set of $n$ bounding boxes denoting a few selected objects, LOCA predicts a density map $\boldsymbol{R} \in \mathbb{R}^{H_0 \times W_0}$ whose values sum into the number of all objects of the selected class present in $\mathbf{I}$.

The LOCA architecture (Figure~\ref{fig:architecture}) follows four steps: 
(i) image feature extraction (encoder), 
(ii) object prototype extraction, 
(iii) prototype matching and (iv) density regression (decoder).
The input image is resized to $H_{IN}\times W_{IN}$ pixels and encoded by a ResNet-50~\cite{resnet} backbone. 
Multi-scale features are extracted from the second, third and fourth block, resized to a common size of $h\times w$ and reduced by $1 \times 1$ convolutional layer into $d$ channels.
To further consolidate the encoded features and increase the similarity between same-category objects, a global (image-wide) self-attention block~\cite{transformer,detr} is applied, thus producing the encoded image features
$\mathbf{f}^{E} \in \mathbb{R}^{h\times w\times d}$.

Next, $n$ object prototypes $\{ \mathbf{q}_{i}^{O} \in \mathbb{R}^{s\times s \times d} \}_{i=1:n}$ with spatial size $s \times s$, corresponding to the annotated bounding boxes are computed by the \textit{object prototype extraction module}, which considers the annotated objects shape and appearance properties (detailed in Section~\ref{sec:object_queries}). The image features $\mathbf{f}^{E}$ are depth-wise correlated with the prototypes. Each prototype thus generates a multi-channel similarity tensor $\tilde{\boldsymbol{R}}_i$, i.e.,
\begin{equation}  \label{eq:correlation}
    \tilde{\boldsymbol{R}}_i = \mathbf{f}^{E} * \mathbf{q}_{i}^{O},
\end{equation}
where $(*)$ is a depth-wise correlation. The individual $n$ prototype similarity tensors are fused by a per-channel, per-pixel max operation, yielding a joint response tensor
$\tilde{\boldsymbol{R}} \in \mathbb{R}^{h\times w\times d}$. 

Finally, a regression head 
predicts the final 2D density map $\boldsymbol{R} \in \mathbb{R}^{H_{IN}\times W_{IN}}$. The regression head consists of three $3\times 3$ convolutional layers with 128, 64 and 32 feature channels, each followed by a Leaky ReLU, a $2\times$ bilinear upsampling layer, and a linear $1 \times 1$ convolution layer followed by a Leaky ReLU. The number of objects in the image $N$ is estimated by summing the density map values, i.e., $N = \mathrm{sum}(\boldsymbol{R})$.

\subsection{Object prototype extraction module}\label{sec:object_queries}

\begin{figure}[!h]
    \centering
    \includegraphics[width=1.0\linewidth]{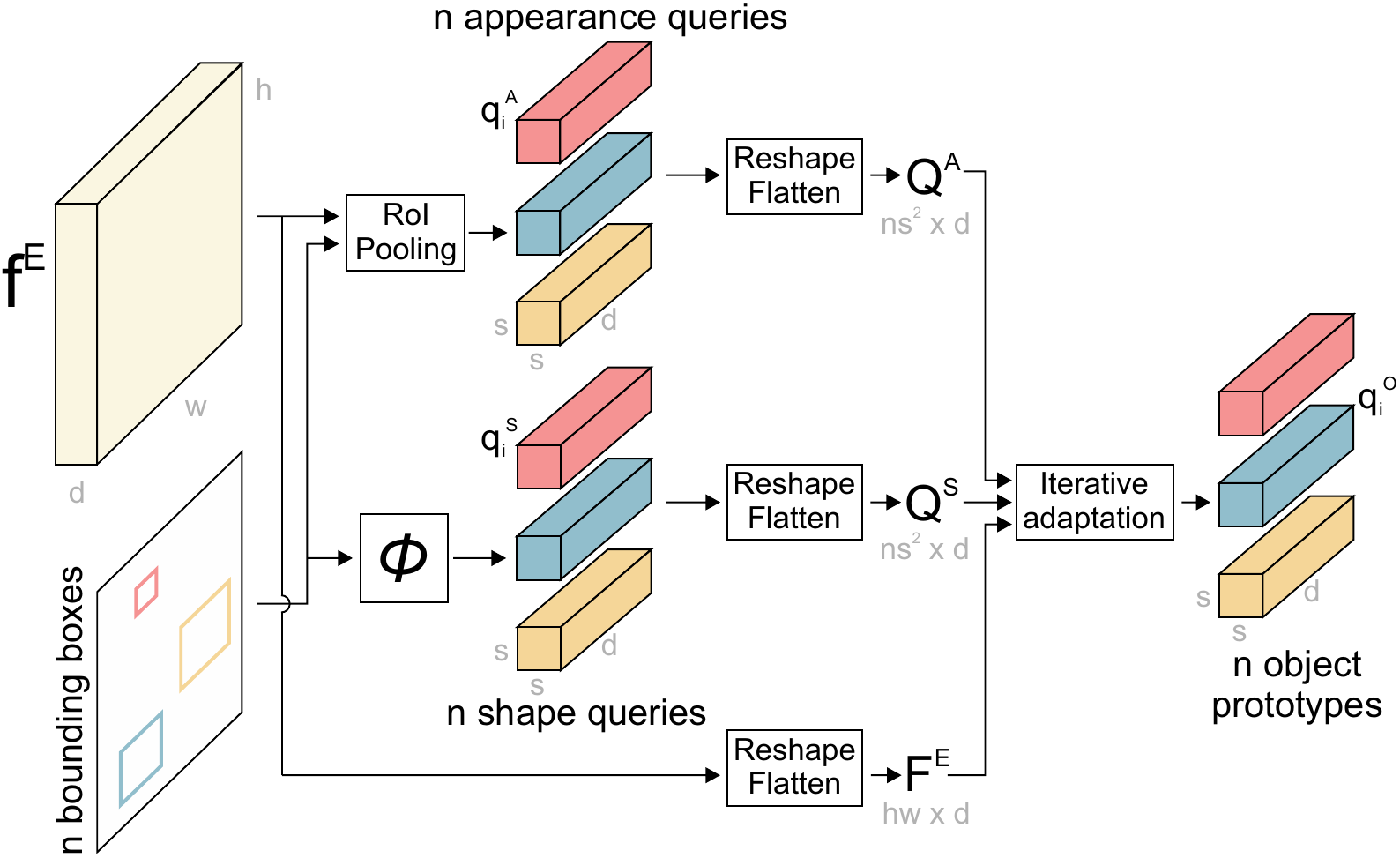}
    \caption{Object prototype extraction module (OPE). Shape and appearance queries are extracted separately and iteratively adapted considering the image-wide information into $n$ object prototypes.
    }
    \label{fig:objectness}
\end{figure}

The object prototype extraction module (OPE) (Figure~\ref{fig:objectness}) constructs $n$ object prototypes 
$\{ \mathbf{q}_{i}^{O} \}_{i=1:n}$, with $\mathbf{q}_{i}^{O} \in \mathbb{R}^{s\times s \times d}$, using the image feature map $\mathbf{f}^{E} \in \mathbb{R}^{h\times w\times d}$ and the set of $n$ bounding boxes $\{ b_i \}_{i=1:n}$. 
Ideally, the prototypes should generalize over the appearance of the selected object category in the image and retain good localization properties. 
Shape information is injected by initializing the prototypes with exemplar width and height features. The appearance of the remaining objects is then iteratively transferred into the final prototypes, with the exemplar appearance supervising the process. We details this process next.

First, $n$ appearance queries $\mathbf{q}_{i}^{A} \in \mathbb{R}^{s\times s \times d}$ are extracted from the annotated objects by RoI pooling~\cite{maskrcnn} the image features $\mathbf{f}^{E}$ from individual bounding boxes $b_i$ into  $s \times s$ tensors.
The pooling operation makes the appearance queries shape-agnostic, since it maps features from different spatial shapes into rectangular queries of the same size.
We introduce shape queries $\mathbf{q}_{i}^{S}$ to recover the lost information as follows. 

The shape query corresponding to the $i$-th bounding box is computed by a nonlinear mapping $\mathbb{R}^2 \rightarrow \mathbb{R}^{s \times s \times d}$ of its width  and height $[b_i^w, b_i^h]$ into a high-dimensional tensor $\mathbf{q}_{i}^{S} = \phi([b_i^w, b_i^h])$. The mapping $\phi(\cdot)$ is implemented as a three-layer feed-forward network ($2 \rightarrow 64 \rightarrow d \rightarrow s^2 d$) with ReLU activations following each linear layer.

\begin{figure}[!t]
    \centering
    \includegraphics[width=1.0\linewidth]{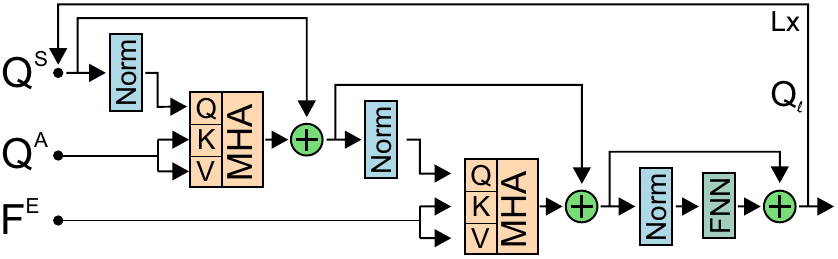}
    \caption{The iterative adaptation module applies attention to gradually generalize prototypes to the object instances indicated by few input exemplars.}
    \label{fig:fusion}
\end{figure}

The shape and appearance queries are converted into object prototypes by an iterative adaptation module (Figure~\ref{fig:fusion}) using a recursive sequence of cross-attention blocks. 
Specifically, the shape queries $\mathbf{q}_{i}^{S}$ are reshaped into a matrix $\mathbf{Q}^{S} \in \mathbb{R}^{n s^2 \times d}$ and in the same way the appearance queries $\mathbf{q}_{i}^{A}$ and image features $\mathbf{f}^{E}$ are reshaped into $\mathbf{Q}^{A} \in \mathbb{R}^{n s^2 \times d}$ and $\mathbf{F}^{E} \in \mathbb{R}^{hw \times d}$, respectively. The adaptation iteration then follows the sequence
\begin{align}
    \mathbf{Q}_\ell' & = \text{MHA}(\text{LN}(\mathbf{Q}_{\ell-1}), \mathbf{Q}^{A},  \mathbf{Q}^{A}) + \mathbf{Q}_{\ell-1} \label{eq:fusion1}\\
    \mathbf{Q}_\ell'' & = \text{MHA}(\text{LN}(\mathbf{Q}_\ell'), \mathbf{F}^{E},  \mathbf{F}^{E}) + \mathbf{Q}_\ell' \label{eq:fusion2}\\
    \mathbf{Q}_\ell & = \text{FFN}(\text{LN}(\mathbf{Q}_\ell'')) + \mathbf{Q}_\ell''\label{eq:fusion3},
\end{align}
where the inputs at $\ell = 0$ are initialized by the shape queries (i.e., $\mathbf{Q}_0 = \mathbf{Q}^{S}$), MHA is the standard multi-head attention~\cite{transformer}, LN is layer normalization and FFN is a small feed-forward network. The process is performed for $L$ iterations, i.e., $\ell \in \{1,...,L\}$.
The output $\mathbf{Q}_L \in \mathbb{R}^{n s^2 \times d}$ is finally reshaped into a set of $n$ object prototypes $\mathbf{q}_{i}^{O} \in \mathbb{R}^{s\times s\times d}$.

\subsubsection{Adaptation for zero-shot setup}\label{sec:zeroshot}

In the zero-shot scenario, the annotation-specific shape and appearance queries cannot be extracted due to absence of object annotations. Thus a minor modification of the OPE module is required to compute the object prototypes $\mathbf{q}_{i}^{O}$. In particular, the step (\ref{eq:fusion1}) is skipped, and $\mathbf{Q}'_\ell$ is initialized by trainable objectness queries $\mathbf{q}_{i}^{S'} \in \mathbb{R}^{s\times s \times d}$. The iterative adaptation module computational sequence then becomes (\ref{eq:fusion2}) and (\ref{eq:fusion3}).

\subsection{Training loss}  \label{sec:loss}

LOCA is trained using the $\ell_2$ loss between the predicted density map $\boldsymbol{R}$ and the ground-truth map $\boldsymbol{\hat{G}}$ normalized by the number of objects,
\begin{equation}
    \label{eq:ose}
    \mathcal{L}_{OSE} = \frac{1}{M}||\boldsymbol{\hat{G}} - \boldsymbol{R}||_2^2,
\end{equation}
where $M$ is the number of objects in the mini-batch.
The normalized loss emphasizes the errors in images with many objects, which usually contain the most challenging situations with high local object densities. 

Auxiliary losses are added to better supervise the training of the iterative adaptation module (Figure~\ref{fig:fusion}). 
In particular, every intermediate output $\mathbf{Q}_\ell$ is reshaped into $n$ queries $\{ \mathbf{q}_i^{\ell} \in \mathbb{R}^{s\times s \times d} \}_{i=1:n}$ and applied to image features $\mathbf{f}^{E}$ as in~(\ref{eq:correlation}), generating an intermediate multi-channel response tensor $\tilde{\boldsymbol{R}}_i^{\ell}$. This is followed by the max operation and regression head to obtain an intermediate density map $\boldsymbol{R}^{\ell}$. The auxiliary loss is then computed as
\begin{equation}
    \label{eq:aux_loss}
    \mathcal{L}_{AUX} =  \frac{1}{M} \sum_{\ell=1}^{L-1} ||\boldsymbol{\hat{G}} - \boldsymbol{R}^{\ell}||_2^2.
\end{equation}
The final loss is thus $\mathcal{L} = \mathcal{L}_{OSE} + \lambda_{AUX} \mathcal{L}_{AUX}$, where $\lambda_{AUX}$ is the auxiliary loss weight.

\section{Experiments}

\subsection{Implementation details}


\textbf{Architecture details.} 
LOCA resizes the input image to $H_{IN}=W_{IN}=512$ pixels and applies the SwAV~\cite{swav} pretrained ResNet50 backbone with the features from the final three blocks upsampled to $h=w=64$ pixels. This results in an activation map with 3584 channels, which is further projected into $d=256$ channels by a $1 \times 1$ convolutional layer. The global self-attention block is a transformer encoder~\cite{detr, transformer} with 3 layers. MHA modules consist of 8 attention heads with the hidden dimension $d=256$, while the FFN has the hidden dimension of 1024. Dropout~\cite{dropout} is applied after every MHA and FFN module with probability 0.1. The iterative adaption module contains $L=3$ layers with the same MHA and FFN dimensions. The object prototype spatial size is $s \times s$ with $s=3$, while the dropout is not used. The ground truth density maps are generated by placing unit densities on object locations and smoothing with the Gaussian kernel, whose size is determined for each image separately. In particular, the kernel size is determined as $1/8$ of the average exemplar bounding box size.

\textbf{Training details.} Standard training image augmentation is applied, such as 
tiling, horizontal flipping and color jitter~\cite{countr}. The backbone network parameters are frozen, while all other LOCA parameters are trained for 200 epochs using the AdamW~\cite{adamw} optimizer with the fixed learning rate $10^{-4}$ and weight decay $10^{-4}$. The auxiliary loss weight in (\ref{eq:aux_loss}) is set to $\lambda_{AUX}=0.3$ and gradient clipping with maximum norm of 0.1 is used. LOCA is trained on two Tesla V100 GPUs with batch size 8 (4 images per GPU) for approximately 10 hours.

\subsection{Comparison with the state of the art}
\label{sec:sota}

LOCA is evaluated on the recent few-shot counting dataset FSC147~\cite{famnet}. The dataset contains 6135 images of 147 object categories split into training, validation and test sets consisting of 3659, 1286 and 1190 images, respectively. The sets of object categories present in each split are disjoint. Each image annotation consists of three bounding boxes of exemplar objects and point annotations for all objects of the same category as the exemplars.


In the few-shot counting scenario, we compare LOCA with GMN~\cite{gmn}, MAML~\cite{maml}, FamNet~\cite{famnet} and the most recent state-of-the-art methods CFOCNet~\cite{cfocnet}, BMNet+~\cite{bmnet}, SAFECount~\cite{safecount} and CounTR~\cite{countr}.  We follow the standard evaluation protocol~\cite{famnet, bmnet, safecount} and compute Mean Absolute Error (MAE) and Root of Mean Squared Error (RMSE) given the predicted and ground truth object counts.  

Results are summarized in Table~\ref{tab:results_fs}. LOCA substantially outperforms all methods with a relative improvement of 22.0~\%, 9.7~\% in terms of MAE on validation and test sets, respectively, and 31.0~\% and 33.4~\% in terms of RMSE and sets a solid new state-of-the-art. Note that LOCA significantly outperforms even the most recent CounTR~\cite{countr}, which applies post-hoc error compensation routines (i.e., it estimates a correction factor for adjusting the estimated count).

\begin{table}[htbp]
    \centering
    \begin{tabular}{l l l l l}
        \toprule
        \multirow{2}{*}{Method}& \multicolumn{2}{c}{Validation set} & \multicolumn{2}{c}{Test set} \\
        & MAE & RMSE & MAE & RMSE \\ 
        \midrule
        GMN~\cite{gmnclass} & 29.66 & 89.81 & 26.52 & 124.57 \\
        MAML~\cite{maml} & 25.54 & 79.44 & 24.90 & 112.68 \\
        FamNet~\cite{famnet} & 23.75 & 69.07 & 22.08 & 99.54 \\
        CFOCNet~\cite{cfocnet} & 21.19 & 61.41 & 22.10 & 112.71 \\        
        BMNet+~\cite{bmnet} & 15.74 & 58.53 & 14.62 & 91.83 \\
        SAFECount~\cite{safecount} & \third{15.28} & \second{47.20} & \third{14.32} & \second{85.54} \\
        CounTR~\cite{countr} & \second{13.13} & \third{49.83} & \second{11.95} & \third{91.23} \\
        LOCA (ours) & \first{10.24} & \first{32.56} & \first{10.79} & \first{56.97} \\
        \bottomrule
    \end{tabular}
    \caption{Evaluation on a few-shot counting scenario.}
    \label{tab:results_fs}
\end{table}

\begin{figure}
    \centering
    \includegraphics[width=\linewidth]{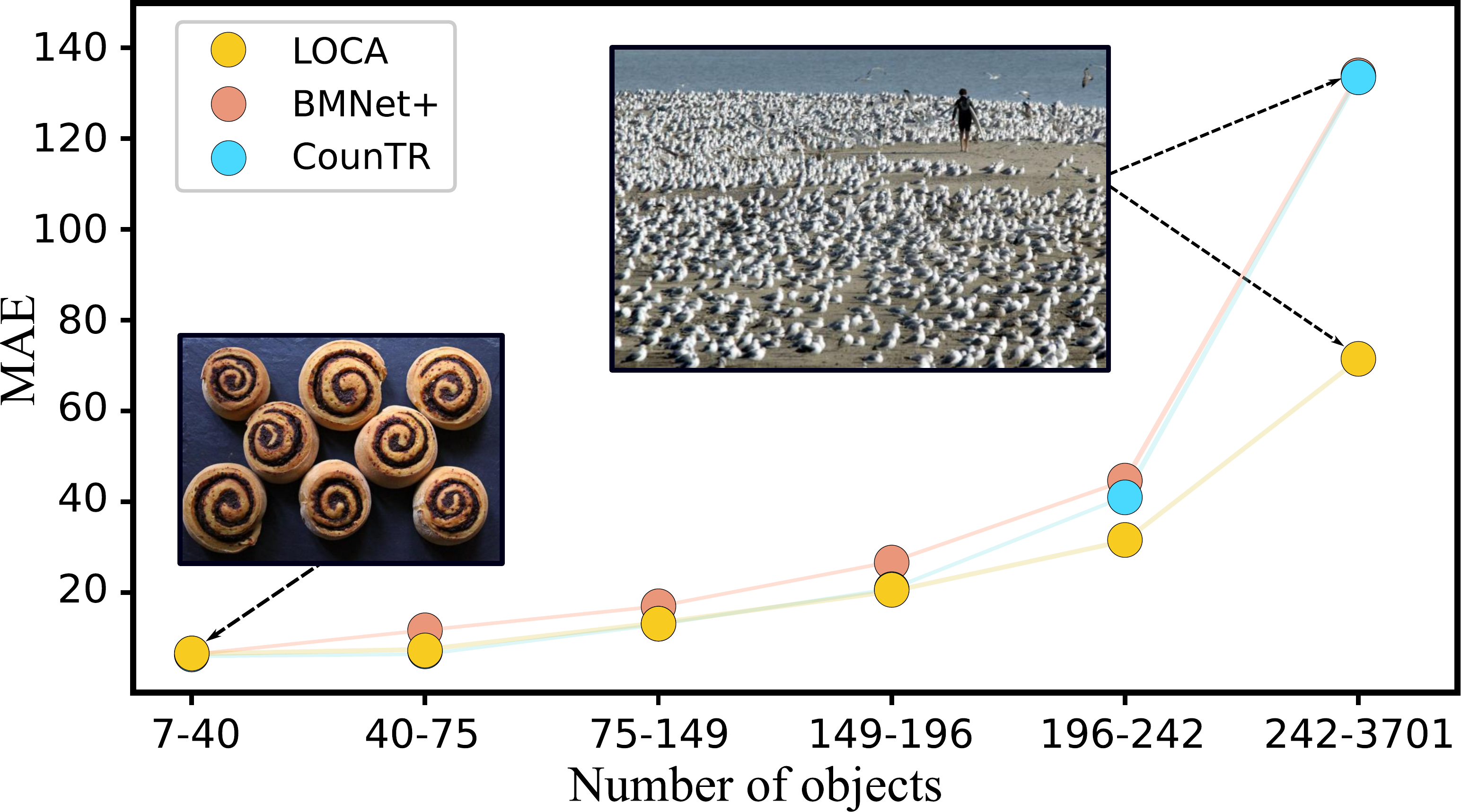}
    \caption{LOCA excells most in the highly challenging dense scenarios and outperforms state-of-the-art accross the density levels.
     }
    \label{fig:obj_count}
\end{figure}

For further insights we inspect the count errors with respect to the number of objects in the image (Figure \ref{fig:obj_count}).
LOCA outperforms the state-of-the-art across the different object numbers and most significantly outperforms the state-of-the-art on images with very high object counts. These typically contain extremely high object densities, presenting substantial challenge to all previous methods. But LOCA copes very well even with these cases, reducing the count errors by nearly $50\%$ compared to state-of-the-art.


\subsubsection{Evaluation on one-shot counting}
We inspect performance under minimal user supervision with a single annotation -- a one-shot scenario. LOCA is compared with LaoNet~\cite{laonet}, which is designed specifically for one-shot scenarios,
as well as with the recent methods GMN~\cite{gmn}, CFOCNet~\cite{cfocnet}, FamNet~\cite{famnet}, BMNet+~\cite{bmnet} and CounTR~\cite{countr} which were specialized for the one-shot setting.
The results are shown in Table~\ref{tab:results_os}. LOCA outperforms the current state-of-the-art with a relative improvement of 13.6~\% MAE on the validation set, and $23.5\%$ and $16.3\%$ RMSE on validation and test set, respectively. This empirically confirms that LOCA generalizes well also to the minimal supervision counting case.

\begin{table}[htbp]
    \centering
    \begin{tabular}{l l l l l}
        \toprule
        \multirow{2}{*}{Method}& \multicolumn{2}{c}{Validation set} & \multicolumn{2}{c}{Test set} \\
        & MAE & RMSE & MAE & RMSE \\ 
        \midrule
        GMN~\cite{gmnclass} & 29.66 & 89.81 & 26.52 & 124.57 \\
        CFOCNet~\cite{cfocnet} & 27.82 & 71.99 & 28.60 & 123.96 \\ 
        FamNet~\cite{famnet} & 26.55 & 77.01 & 26.76 & 110.95 \\
        BMNet+~\cite{bmnet} & 17.89 & 61.12 & 16.89 & \third{96.65} \\
        LaoNet~\cite{laonet}& \third{17.11} & \third{56.81} & \third{15.78} & 97.15 \\
        CounTR~\cite{countr}& \second{13.15} & \second{49.72} & \first{12.06} & \second{90.01} \\
        LOCA (ours) & \first{11.36} & \first{38.04} & \second{12.53} & \first{75.32} \\
        \bottomrule
    \end{tabular}
    \caption{Evaluation on a one-shot counting scenario.}
    \label{tab:results_os}
\end{table}

\subsubsection{Evaluation on zero-shot counting}


As noted in Section~\ref{sec:zeroshot}, LOCA can be easily applied to the unsupervised counting scenario with no user annotations, i.e., the zero-shot setup. 
We thus compare LOCA with zero-shot CounTR~\cite{countr} and state-of-the-art methods RepRPN-C~\cite{reprpn} and RCC~\cite{hobley} which are specialized for zero-shot counting. 
The results in Table~\ref{tab:results_zs} show that LOCA achieves relative improvements of 6.5~\%, and 0.5~\% in terms of RMSE on validation and test sets, respectively, compared to the state-of-the-art, and outperforms all zero-shot specialized architectures. This confirms that the proposed OPE module successfully adapts the trainable objectness queries into strong object prototypes capable of accurate count estimation even in the extreme case without manually annotated exemplars.


\begin{table}[htbp]
    \centering
    \begin{tabular}{l l l l l}
        \toprule
        \multirow{2}{*}{Method}& \multicolumn{2}{c}{Validation set} & \multicolumn{2}{c}{Test set} \\
        & MAE & RMSE & MAE & RMSE \\ 
        \midrule
        RepRPN-C \cite{reprpn} & 29.24 & 98.11 & 26.66 & 129.11 \\
        RCC \cite{hobley} & \third{17.49} & \second{58.81} & \third{17.12} & \second{104.53} \\
        CounTR \cite{countr} & \first{17.40} & \third{70.33} & \first{14.12} & \third{108.01} \\
        LOCA (ours) & \second{17.43} & \first{54.96} & \second{16.22} & \first{103.96} \\
        \bottomrule
    \end{tabular}
    \caption{Evaluation on a zero-shot counting scenario.}
    \label{tab:results_zs}
\end{table}

\subsubsection{Qualitative few-shot counting results}

\begin{figure*}
    \centering
    \includegraphics[width=\linewidth]{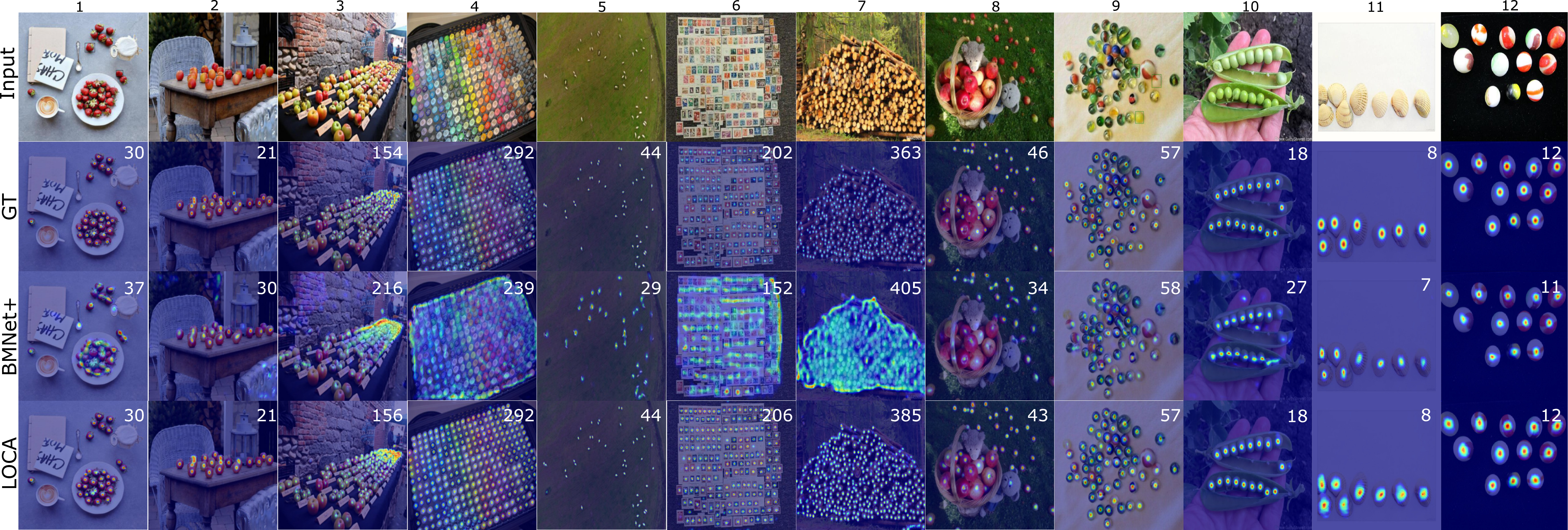}
    \caption{Qualitative results on the FSC147 dataset. Compared to related works, LOCA better discriminates between objects and background, predicts density maps with clear object locations and works well on smaller objects, while better capturing the intra-class variability and within-image size variability.}
    \label{fig:sota}
\end{figure*}

Figure~\ref{fig:sota} visualizes the predicted object density maps from LOCA and BMNet+~\cite{bmnet}. Note that LOCA produces density maps with high fidelity object localization. The prototypes generated by the OPE module discriminate well between the objects and the background (columns 1--3). In the second column, LOCA generates clear density peaks on the object centers. In columns 4, 5, 6 and 7 we see that LOCA outperforms BMNet+ on small objects, with objects localized far better in the density map. This is likely due to explicitly accounting for object shape and size by shape-specific objectness queries, in contrast to other methods that consider only scale-agnostic object appearance extraction.
The shape-specific information enables LOCA to more robustly address the object size variability within the image. In column 8, BMNet+ misses several larger apples while LOCA accurately localizes apples of all sizes. Similarly, BMNet+ underestimates the density of larger marbles in column 9, while LOCA produces a much more crisp and accurate density map. Columns 10, 11 and 12 show examples with few objects. LOCA also performs well in such scenarios.




\subsection{Comparison with object detectors} 
\label{sec:obj_det}

In limited cases where large training sets are available, objects can be counted using pretrained object detectors.
It is thus instructive to evaluate the general few-shot object counters in these specialized cases in comparison with the classical detectors. The FSC147 dataset~\cite{famnet} in fact provides image subsets Val-COCO and Test-COCO containing only categories for which abundant annotated training images are available in COCO~\cite{cocodata}. 

This allows comparing counting capabilities of LOCA with those of classical detectors FasterRCNN~\cite{fasterrcnn}, MaskRCNN~\cite{maskrcnn}, and RetinaNet~\cite{retinanet} as well as the recent few-shot counting state-of-the-art FamNet~\cite{famnet}, BMNet+~\cite{bmnet} and CounTR~\cite{countr}. The results are reported in Table~\ref{tab:results_coco}. LOCA achieves state-of-the-art performance, most significantly on Val-COCO with a relative $31\%$ MAE and $36\%$ RMSE improvement over the best method.


\begin{table}[htbp]
    \centering
    \begin{tabular}{l l l l l}
        \toprule
        \multirow{2}{*}{Method}& \multicolumn{2}{c}{Val-COCO} & \multicolumn{2}{c}{Test-COCO} \\
        & MAE & RMSE & MAE & RMSE \\ 
        \midrule
        Faster-RCNN \cite{fasterrcnn} & 52.79 & 172.46 & 36.20 & 79.59 \\
        RetinaNet \cite{retinanet} & 63.57 & 174.36 & 52.67 & 85.86 \\
        Mask-RCNN \cite{maskrcnn} & 52.51 & 172.21 & 35.56 & 80.00 \\
        Famnet \cite{famnet} & 39.82 & 108.13 & 22.76 & 45.92 \\
        BMNet+ \cite{bmnet} & \third{26.55} & \third{93.63} & \third{12.38} & \first{24.76} \\
        CounTR \cite{countr} & \second{24.66} & \second{83.84} & \second{10.89} & \second{31.11} \\
        LOCA (ours) & \first{16.86} & \first{53.22} & \first{10.73} & \third{31.31} \\
        \bottomrule
    \end{tabular}
    \caption{Evaluation on the COCO object-detection counting dataset.}
    \label{tab:results_coco}
\end{table}

\subsection{Cross-dataset generalization}
\label{sec:cross_dataset}

We evaluate the cross-dataset generalization capabilities of LOCA using the established evaluation protocol from~\cite{famnet}. In that protocol, a method is trained on the FSC147 dataset~\cite{famnet} and evaluated on the CARPK dataset~\cite{carpk}, which is a car-counting dataset containing aerial images of parking lots, which are considerably different from the FSC147 images. To ensure there is no object class overlap between the training and test dataset, the car images are omitted from the FSC-147 training set. For counting purposes, twelve exemplars are sampled from the training CARPK images and used in all test CARPK images. 


The results are reported in Table~\ref{tab:cars}. LOCA achieves better cross-dataset generalization with a relative  4.5~\% MAE and 9.2~\% RMSE improvement compared to the most recently published state-of-the-art method BMNet+, thus setting a new dataset generalization state-of-the-art among the few-shot counting methods.

\begin{table}[htbp]
    \centering
    \begin{tabular}{c c c}
        \toprule
        Method & MAE & RMSE \\ 
        \midrule
        FamNet~\cite{famnet} & \third{28.84} & \third{44.47} \\
        BMNet+~\cite{bmnet} & \second{10.44} & \second{13.77} \\
        LOCA (ours) & \first{9.97} & \first{12.51} \\
        \bottomrule
    \end{tabular}
    \caption{Cross-dataset generalization experiment on CARPK~\cite{carpk}. 
    }
    \label{tab:cars}
\end{table}

\subsection{Ablation study}
\label{sec:ablation}


We finally analyze the architectural design choices and examine the influence of the object-normalized loss and the auxiliary losses. 
The experiments are performed on the FSC147 dataset in the few-shot setting. 
We report the performance by averaging a certain measure on validation and test sets. 


\noindent{\bf Architecture design.} 
Table~\ref{tab:ablation_components} reports the performance of re-trained LOCA variants with individual computational blocks removed. 
To evaluate the importance of global attention in the image features encoder block, we removed this block ($LOCA_\mathrm{no\_att}$) and observe a  $12\%$ MAE performance drop. This indicates the importance of image feature consolidation by attention, which likely brings objects of the same category closer at feature level. 
The impact of the OPE module is evaluated by removing it and extracting the object prototypes directly from the encoder image features by pooling the features of the exemplar regions ($LOCA_\mathrm{no\_ope}$). This results in significant performance drop in order of $34\%$ MAE.
Removing both, global attention and OPE ($LOCA_\mathrm{no\_att\_ope}$) leads to further performance drops of $\sim 39\%$ MAE compared to the original LOCA.

Next, we explored the importance of the exemplar shape information in addition to their appearance. A variant $LOCA_\mathrm{no\_shape}$ was constructed, which ignores the shape queries $\mathbf{q}_i^S $ by omitting the first attention block in the OPE module (Figure~\ref{fig:objectness}) and replacing 
$Q^S$ with $Q^A$ in the second attention block. We observe a $25\%$ MAE reduction compared to original LOCA. This confirms the importance of accounting for the shape information in addition to the appearance in OPE.

Importance of the mapping function that transforms the exemplar width and height into shape-specific objectness queries (Section~\ref{sec:object_queries}) is analyzed in the following. 
Instead of predicting the shape queries from exemplars, we replace them with trainable queries ($LOCA_\mathrm{pre\_shape}$). 
Results show a significant drop in performance compared to LOCA with a $27\%$ change in MAE, indicating that useful shape-specific objectness information is indeed extracted from the exemplars size parameters and that it significantly contributes to object localization and accurate counts.

We also analyzed the role of the first cross-attention in the first OPE iteration (Equation~\ref{eq:fusion1}) by replacing it with a simple summation:
$\mathbf{Q}_1' = \mathbf{Q}^{A} + \mathbf{Q}^S$.
This results in a $5\%$ increase of MAE and a $22\%$ increase of RMSE, which indicates that the first MHA in OPE should not be considered as a simple matching operation, but rather as a modulation of the prototype construction process by the exemplar shape information.
This information is unique for every exemplar, thus optimally adjusting the resulting prototype to localize the objects of interest. 

Finally, we analyzed the impact of the number of adaptation iterations $L$ in the iterative adaptation module (Section~\ref{sec:object_queries}) on the joined FSC-147 evaluation sets. 
Results are shown in Table~\ref{tab:ablation_L}. 
The choice of $L=3$ provides the best performance while maintaining a low model complexity.

\noindent{\bf Complexity.} As shown in Table~\ref{tab:params}, the proposed architecture has almost $3 \times$ less parameters and almost $10 \times$ less trainable parameters than CounTR while being comparable to other state-of-the-art methods in both the number of parameters and computational complexity.
These results demonstrate that excellent low-shot object counting performance of LOCA comes from the methodological improvements instead of increased complexity.

\begin{table}[htbp]
    \centering
    \begin{tabular}{l c c c c}
        \toprule
        \multirow{2}{*}{Method}& \multicolumn{2}{c}{Validation set} & \multicolumn{2}{c}{Test set} \\
        & MAE & RMSE & MAE & RMSE \\ 
        \midrule
        $LOCA_\mathrm{no\_att\_ope}$  & 17.62 & 55.78 & 16.92 & 101.96 \\
        $LOCA_{no\_ope}$  & 16.24 & 57.41 & 15.53 & 96.23 \\
        $LOCA_{no\_shape}$   & 13.77 & 49.60 & 14.29 & 112.48 \\
        $LOCA_{pre\_shape}$   & 13.00 & 44.61 & 15.80 & 122.87 \\
        $LOCA_{no\_att}$ & 11.99 & 36.67 & 11.96 & 78.72 \\
        LOCA               & 10.24 & 32.56 & 10.79 & 56.97 \\
        \bottomrule
    \end{tabular}
    \caption{Ablation study of individual architectural components. $LOCA_\mathrm{no\_att\_ope}$ removes the entire global self-attention block from the encoder and the OPE module,  $LOCA_{no\_ope}$ removes the OPE module, $LOCA_{no\_shape}$ removes the use of the shape queries and $LOCA_{no\_att}$ removes the global self-attention block from the encoder.}
    \label{tab:ablation_components}
\end{table}


\begin{table}[htbp]
    \centering
    \begin{tabular}{c c c c c c c}
        \toprule
        L & 1 & 2 & 3 & 4 & 5 & 6\\ 
        \midrule
        MAE & 11.07 & 10.81 & 10.50 & 10.89 & 11.60 & 11.04 \\
        \bottomrule
    \end{tabular}
    \caption{Ablation of the number of iterations $L$ in the iterative adaptation module. 
    }
    \label{tab:ablation_L}
\end{table}

\begin{table}[htbp]
\centering
\begin{tabular}{c c c c}
 \toprule
 \multirow{2}{*}{Method} & \multirow{2}{*}{GFLOPS} & \multicolumn{2}{c}{Number of parameters} \\
 & & Total & Trainable \\
 \midrule
 FamNet & 55 & 26M & 760k \\
 BMNet+ & 27 & 13M & 12M \\
 SafeCount & 366 & 32M & 20M  \\
 CounTR & 91 & 100M & 99M \\
 LOCA (ours) & 80 & 37M & 11M \\  
 \bottomrule
\end{tabular}
\caption{Computational complexity and the number of parameters.}
\label{tab:params}
\end{table}

\noindent{\bf Backbone and resolution.} 
Importance of the backbone pre-training regime, input image resolution and the prototype spatial size are presented in Table~\ref{tab:other}.
Replacing the SwAV-pretrained backbone with the ImageNet-pretrained one ($LOCA_{ImNet}$) results in only a slight performance drop (8\% MAE and 4\% RMSE). 
Reducing the input image resolution from $512 \times 512$ to $384 \times 384$ pixels ($LOCA_{384}$) leads to the 9\% performance drop in both MAE and RMSE. 
Without any hyperparameter modifications, $LOCA_{384}$ remains the top-performing method in three out of four metrics.
Changing the prototype spatial size $s$ from 3 to 1 ($LOCA_{s=1}$) or 5 ($LOCA_{s=5}$) does not lead to significant performance drops. 
MAE is reduced by 3\% and 10\% while RMSE is reduced by 6\% and 11\% for $s=1$ and $s=5$, respectively, which confirms that LOCA is not sensitive to the prototype spatial size. 
All these results further verify that the design of the OPE module is the main driver of LOCA's superior performance.

\begin{table}[htbp]
    \centering
    \begin{tabular}{l c c c c}
        \toprule
        \multirow{2}{*}{Method}& \multicolumn{2}{c}{Validation set} & \multicolumn{2}{c}{Test set} \\
        & MAE & RMSE & MAE & RMSE \\ 
        \midrule
        $LOCA_{ImNet}$         & 11.40 & 37.10 & 11.56 & 55.89 \\
        $LOCA_{384}$           & 10.26 & 32.62 & 12.75 & 65.34 \\
        $LOCA_{s=1}$           & 10.90 & 38.66 & 10.79 & 56.97 \\
        $LOCA_{s=5}$           & 11.11 & 35.47 & 12.27 & 65.08 \\
        LOCA                   & 10.24 & 32.56 & 10.79 & 56.97 \\
        \bottomrule
    \end{tabular}
    \caption{Impact of the backbone pre-training regime, input image resolution and the prototype spatial size on LOCA's performance.}
    \label{tab:other}
\end{table}

\noindent{\bf Model supervision.} We explored the impact of object count normalization in $\mathcal{L}_{OSE}$ (Equation~\ref{eq:ose}) and the importance of using the auxiliary losses on OPE blocks (Section~\ref{sec:loss}). 
Results are shown in Table~\ref{tab:ablation_loss_function}. Avoiding the object count normalization leads to a $11\%$ performance drop in terms of MAE. This shows the benefits of the object count normalization which places a larger penalty on images with larger object counts providing an emphasis on difficult cases with high local object densities.
Additionally removing the auxiliary losses leads to a $17\%$ performance drop in terms of RMSE. This drop in performance indicates that supervision on individual iterations in OPE is beneficial as it encourages the OPE module to provide informative features throughout the iterative process.

\begin{table}[htbp]
    \centering
    \resizebox{1.0 \linewidth}{!}{\begin{tabular}{c c c c c c}
        \toprule
        \multirow{2}{*}{$\mathcal{L}_{OSE}$} & \multirow{2}{*}{Auxiliary loss} & \multicolumn{2}{c}{Validation set} & \multicolumn{2}{c}{Test set} \\
        & & MAE & RMSE & MAE & RMSE \\ 
        \midrule
        \xmark & \xmark & 10.87 & 35.68 & 11.93 & 72.83 \\
        \xmark & \cmark & 10.86 & 31.89 & 12.83 & 62.73 \\
        \cmark & \cmark & 10.24 & 32.56 & 10.79 & 56.97 \\
        \bottomrule
    \end{tabular}}
    \caption{Ablation study on the object-normalized $\ell_2$ loss ($\mathcal{L}_{OSE}$) and the auxiliary losses after every block in OPE. The mark \xmark with $\mathcal{L}_{OSE}$ indicates that the standard $\ell_2$~\cite{famnet} loss is used.}
    \label{tab:ablation_loss_function}
\end{table}

\subsection{Qualitative analysis}

\begin{figure}
    \centering
    \includegraphics[width=\linewidth]{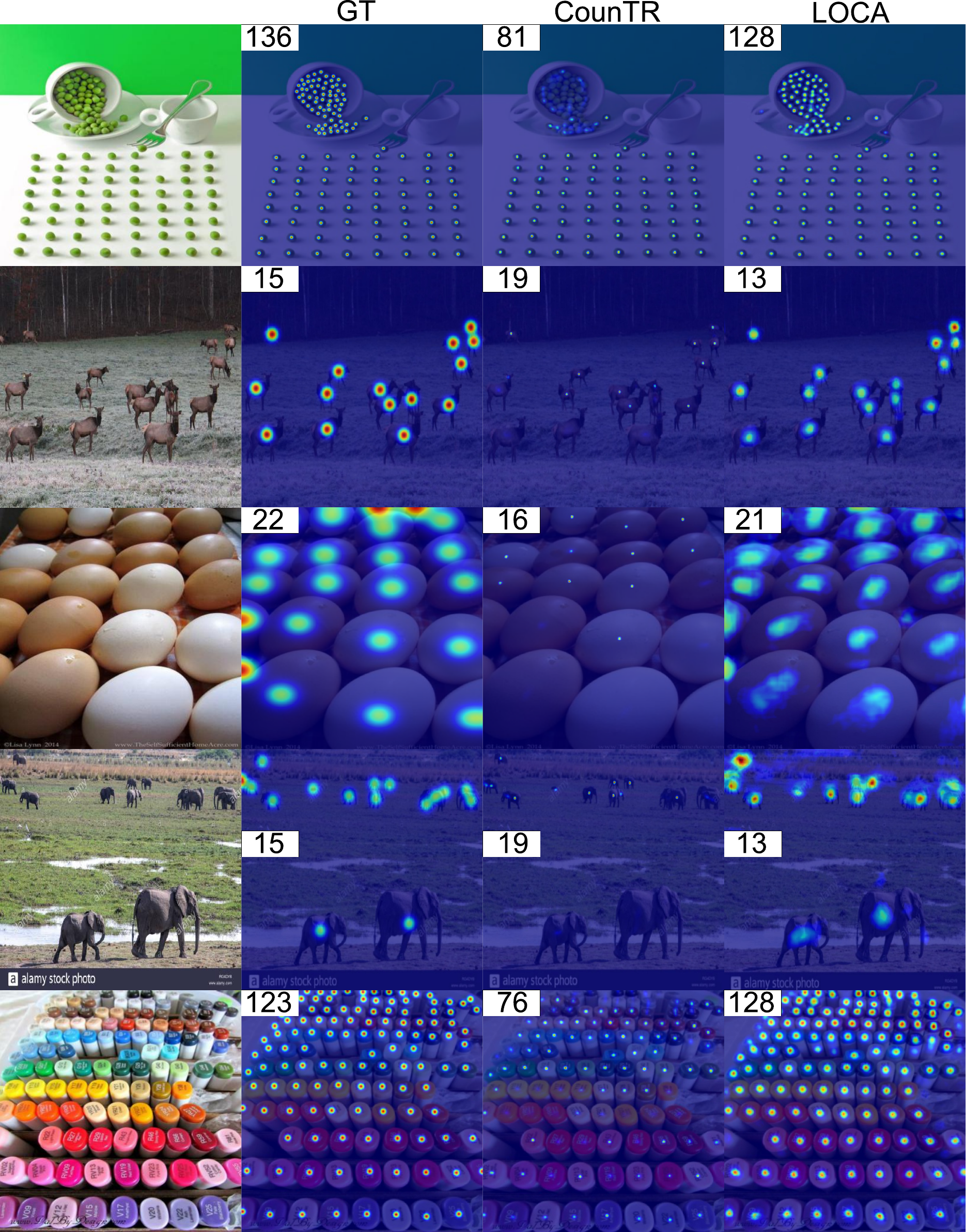}
    \caption{LOCA demonstrates superior performance on images with only small objects (first and second row), images with only large objects (third row), as well as images with objects of mixed sizes (fourth and fifth row).}
    \label{fig:loca_vs_countr}
\end{figure}

\begin{figure}
    \centering
    \includegraphics[width=\linewidth]{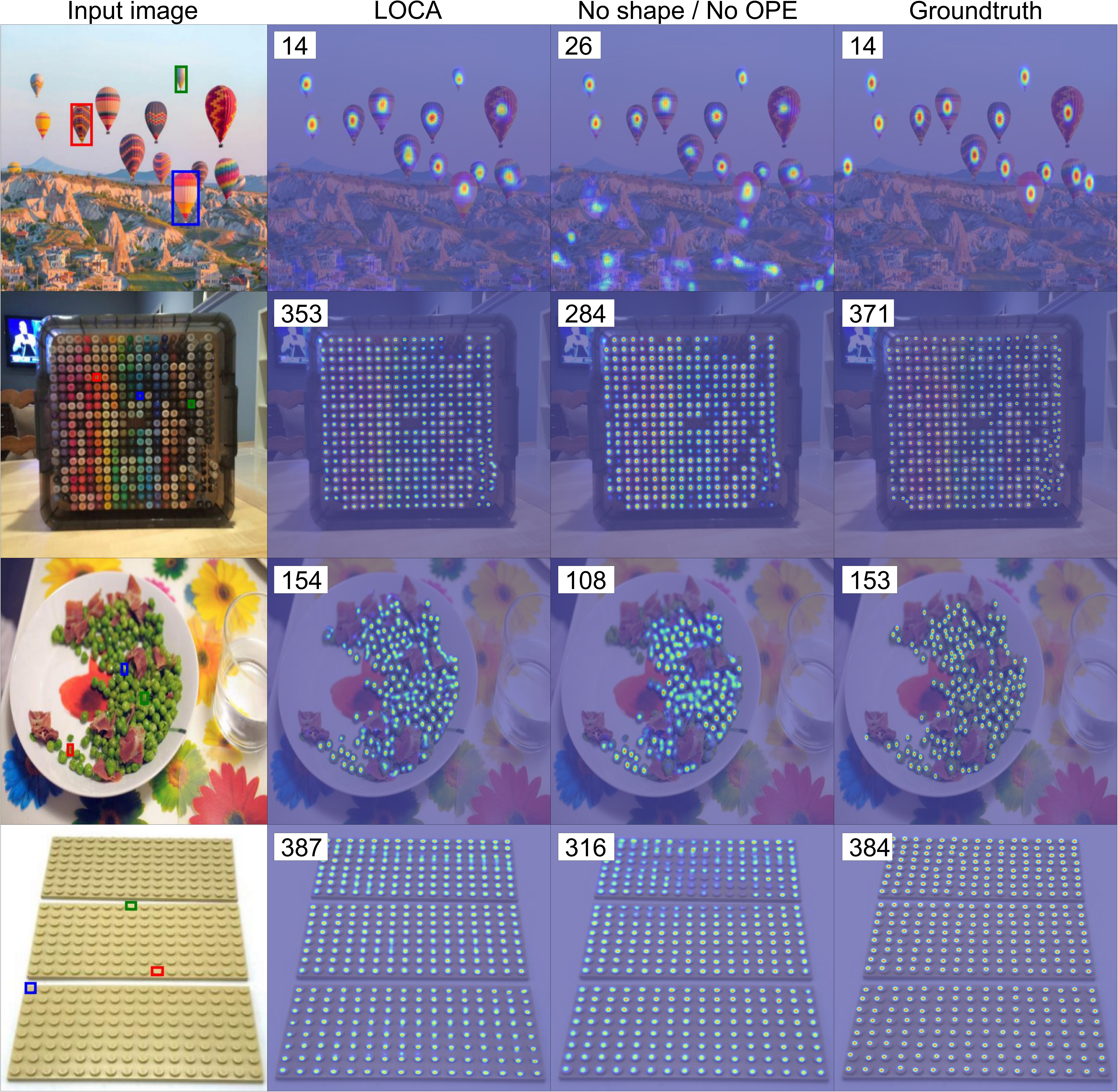}
    \caption{
    LOCA compared to a variant that does not use shape information (first two rows) and a variant without the OPE module (last two rows). Objects across scales are most accurately localized and counted when using the shape information and OPE. 
    }
    \label{fig:loca_vs_modified}
\end{figure}

Figure~\ref{fig:loca_vs_countr} qualitatively compares LOCA with the recent 
state-of-the-art method CounTR~\cite{countr}. LOCA demonstrates superior performance in counting small objects (first and second row), large objects (third row) and objects of mixed sizes (fourth and fifth row), which supports the proposed design. Figure~\ref{fig:loca_vs_modified} qualitatively compares LOCA with a version that does not use shape information and a version without the OPE module. The shape information injection and the adaptation in OPE module both contribute to accurate localization and counts.


\section{Conclusion}

We presented a new low-shot counting method LOCA, that addresses the limitations of the current state-of-the-art methods. LOCA considers the exemplar shape and appearance properties separately and iteratively adapts these into object prototypes by a new object prototype extraction (OPE) module considering the image-wide features. The prototypes thus generalize to the non-annotated objects in the image, leading to better localization properties and count estimates. 


Experiments show that LOCA outperforms state-of-the-art on the FSC147 public benchmark in few-shot, one-shot and zero-shot settings. 
We observed 
a relative RMSE improvement of $33.4\%$ in few-shot and $16.3\%$ in zero-shot scenarios.
On the COCO subsets of FSC147, LOCA outperforms recent state-of-the-art counting methods, as well as object detection methods, achieving a $36\%$ RMSE improvement. 
On the CARPK dataset, LOCA achieves a relative improvement of $9.2\%$ RMSE, which demonstrates excellent cross-dataset generalization. 
The quantitative results convincingly support the benefits of the new OPE module, which is our main contribution.

We envision several possible future research directions. 
Additional supervision levels such as introducing negative exemplar annotations could be introduced in LOCA for better specification of the selected object class. 
This could lead to interactive tools for accurate object counting. Furthermore, a gap between low-shot counters and object detectors could be further narrowed by enabling bounding box or segmentation mask prediction in LOCA to output additional statistics about the counted objects such as average size, etc., which is useful for many practical applications such as biomedical analysis.

\hfill 

{\noindent \bf Acknowledgements}:
This work was supported by Slovenian research agency program 
P2-0214 
and projects 
J2-2506, 
Z2-4459, 
23-20MR.R588 
and 
J2-3169. 

{\small
\bibliographystyle{ieee_fullname}
\bibliography{egbib}
}
\end{document}